\documentclass[letterpaper, 10 pt, conference]{ieeeconf}  
\usepackage{amsmath,amsfonts}
\usepackage{cite}
\usepackage{algorithmic}
\usepackage{algorithm}
\usepackage{multirow}
\usepackage{array}
\usepackage[caption=false,font=normalsize,labelfont=sf,textfont=sf]{subfig}
\usepackage{textcomp}
\usepackage{stfloats}
\usepackage{url}
\usepackage{verbatim}
\usepackage{graphicx}
\usepackage{caption}

\usepackage{cleveref}
\usepackage[font=small]{caption}  

 \usepackage{stfloats}

\IEEEoverridecommandlockouts                              

\title{\LARGE \bf
Control the Soft Robot Arm with its Physical Twin
}

\author{Qinghua Guan‡*, Hung Hon Cheng‡, Benhui Dai and Josie Hughes*
\thanks{Qinghua Guan$^{‡*}$ (qinghua.guan@epfl.ch), Hung Hon Cheng$^{‡}$ (hung.cheng@epfl.ch), Benhui Dai (benhui.dai@epfl.ch), Josie Hughes$^{*}$ (josie.hughes@epfl.ch) are with the CREATE Lab, School of Engineering STI, EPFL, Swiss.
*: corresponding author,
‡: These authors contributed equally to this work.}
\thanks{Note: This work has been submitted to the IEEE for possible publication. Copyright may be transferred without notice, after which this version may no longer be accessible.}
}

\begin{document}

\maketitle
\thispagestyle{empty}
\pagestyle{empty}

\begin{abstract}
To exploit the compliant capabilities of soft robot arms we require controller which can exploit their physical capabilities.  Teleoperation, leveraging a human in the loop, is a key step towards achieving more complex control strategies. 
Whilst teleoperation is widely used for rigid robots, for soft robots we require teleoperation methods where the configuration of the whole body is considered.
We propose a method of using an identical 'physical twin', or demonstrator of the robot.  This tendon robot can be back-driven, with the tendon lengths providing configuration perception, and enabling a direct mapping of tendon lengths for the execture. 
We demonstrate how this teleoperation across the entire configuration of the robot enables complex interactions with exploit the envrionment, such as squeezing into gaps. 
We also show how this method can generalize to robots which are a larger scale that the physical twin, and how, tuneability of the stiffness properties of the physical twin simplify its use.

\end{abstract}

\section{Introduction}
 
Soft robotic arms, with their inherent compliance and adaptability, offer advantages in applications requiring safe environmental interactions~\cite{dou2021soft,ghobadi2024beyond,hughes2016soft}.  However, their deformable structures pose significant challenges in whole-arm configuration control compared to rigid robotic arms~\cite{wang2022control}, which benefit from well-established kinematic models and precise motion control frameworks. Unlike rigid arms, where forward and inverse kinematics are well-defined, soft robots exhibit highly nonlinear behavior, making traditional model-based or simulation-driven control approaches computationally expensive and difficult to generalize \cite{ambaye2024soft,relano2022modeling,chen2024data}. An alternative approach is to leverage humans in the loop to demonstrate, control or teleoperate the soft robot arm, removing the reliance on precise kinematic control~\cite{amaya2021evaluation,darvish2023teleoperation}.


\begin{figure}[h!]
\centering
\includegraphics[width=0.45\textwidth]{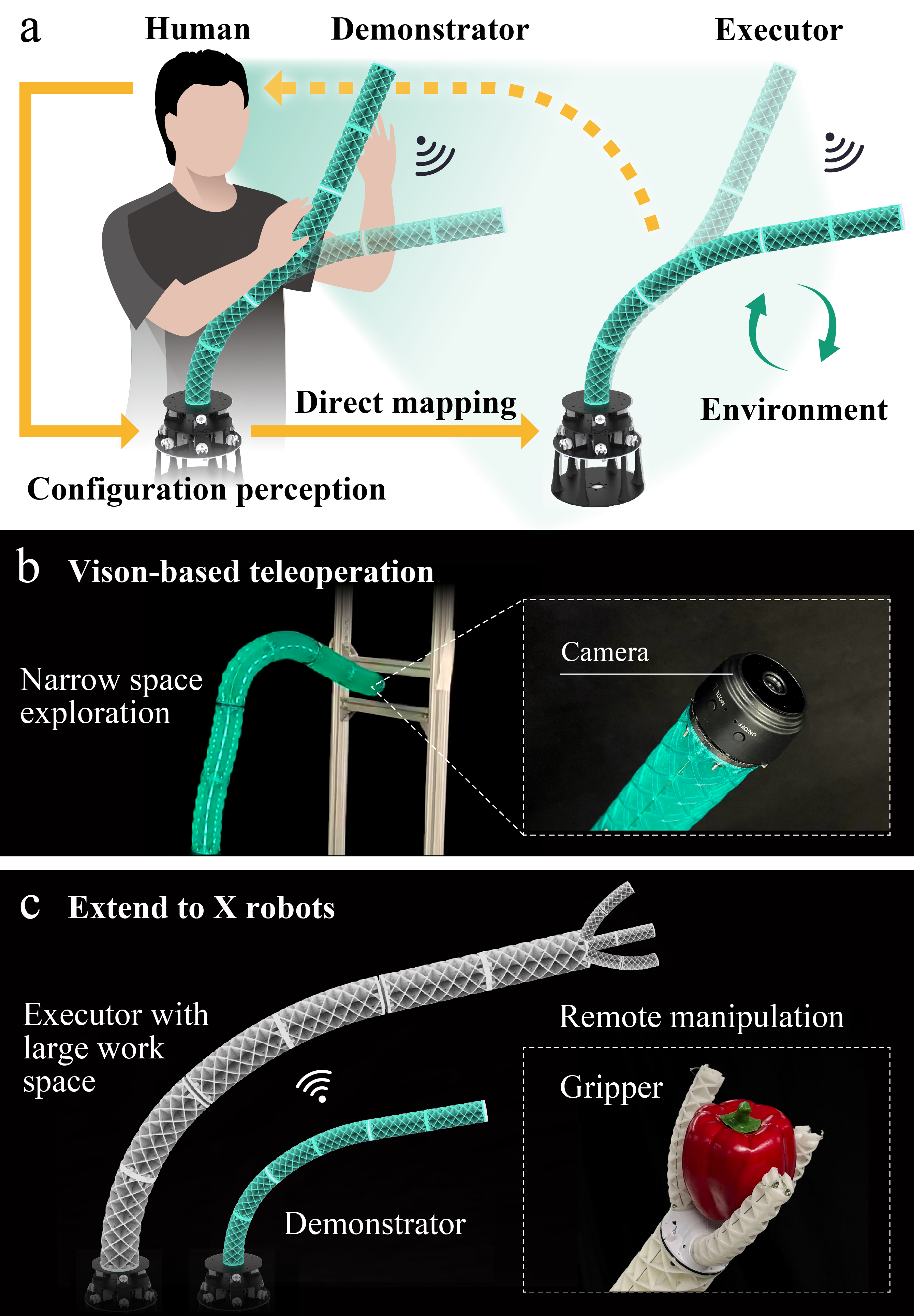}
\captionsetup{justification=centering}
\caption{a) Teleoperation of a soft arm leveraging a physical twin to achieve implicit configuration perception and direct mapping, b) The capabilities and uses of this teleoperation method, and c) The ability to map from demonstrator to robots of different scales (i.e. a 1:X scale mapping) }
\label{fig: Concept physical twin}
\end{figure}

A number of control methods leveraging humans feedback have been demonstrated for soft robot arms.  
One approach is incorporating sensors into the arm that humans can use to control the kinematics of the arm, a notable example being the use of  touch-less proximity sensors~\cite{liu2022touchless,ku2024soft}. Whilst this can enable precise positioning, this requires the human to be directly next to the robot and only allows control of the end effector location of the robot. 
Connecting haptic devices for position control~\cite{uthai2025haptics, amaya2021evaluation}, or leveraging virtual visualization and interactions~\cite{bern2024soft,amaya2021evaluation} can allow for more full configuration space control with feedback, but still typically only enables control of discrete points along the robots body.
Gesture-based interfaces can leverage natural human motion for more intuitive manipulation~\cite{stroppa2020human,lai2024gesture,thakur2024tetherless,phillips2018dexterous}, but there remains challenges in meaningfully mapping gestures to full arm configuration.
As such, there is an unresolved challenge of remote or in-direct control of the entire configuration of soft robot arms via a human interactions. This must be intuitive, fast and enable the human to exploit the compliance of the robot. 




In this work, we propose \textit{Physical Twin Control (PTC)} framework for direct teleoperation of the full configuration of a soft, tendon driven robot arm.  The soft robot arm and the physical twin are identical, such that the physical twin serves as a real-world 'simulation' of the robot it controls.  It has the same configuration space and can naturally demonstrate the same deformation, as shown in Fig.~\ref{fig: Concept physical twin}a.
This allows the tendons in the demonstrator physical twin to implicitly capture its full configuration, and hold this position with admittance control.  Then, with minimal computation effort, this can be mapped to the kinematic control of the soft robot, enabling a mapping with high configuration.  
By leveraging the inherent self-configuration perception of the physical twin, operators can achieve intuitive, global interaction of the entire robotic system, enabling the transfer of complex motions, which can leverage the compliant capabilities of the robot.
In addition to demonstrating the dexterous capabilities which can be achieved with this method, we show how tuning the stiffness and friction properties across the length of the physical twin demonstrator can advantageously alter the region of the soft arm which in under control.

In the remainder of this paper we first introduce the methods for the design and control of the demonstrator and executor robot.  We show the resulting performance, and highlight how online tuning of the stiffness of the physical twin can vary the configurability of the controlled robot.

\section{Methods}

\subsection{Soft Robot Design}

The soft robot arm used both as demonstrator and executer is a tendon-driven system consisting of three flexible sections based on the Trimmed Helicoid (TH) structure, as illustrated in Fig. \ref{fig: Manipulator Design}a~\cite{guan2023trimmed}. 
The TH structure exhibits high axial stiffness while maintaining low bending stiffness, enabling the arm to achieve a higher load capacity and a larger workspace, albeit with some compromise on compliance.
Each sections of the TH robotic arm is actuated by 3 tendons arranged at 120° intervals along the arm, resulting in total 9 tendons. Specifically, the tendon placements for Sections I and III are at 60°, 180°, and 300°, whereas those for Section II are positioned at 120°, 240°, and 0°, as depicted in Fig.~\ref{fig: Manipulator Design}a.

To minimize unintended interactions between sections, bowden tubes guide the cables from the motor base to the starting points of specific sections.  These polytetrafluoroethylene (PTFE) tubes have low friction internally and externally, so there is minimal friction in the cable transmission, and when the bowden tunes move of the surface of the robot. Further details regarding the design can be found in our previous work ~\cite{guan2023trimmed}. 

\subsection{Demonstrator Arm Control: Back-drivable Admittance Control with Proprioception}

\begin{figure*}[htb]
\centering
\includegraphics[width=0.85\textwidth]{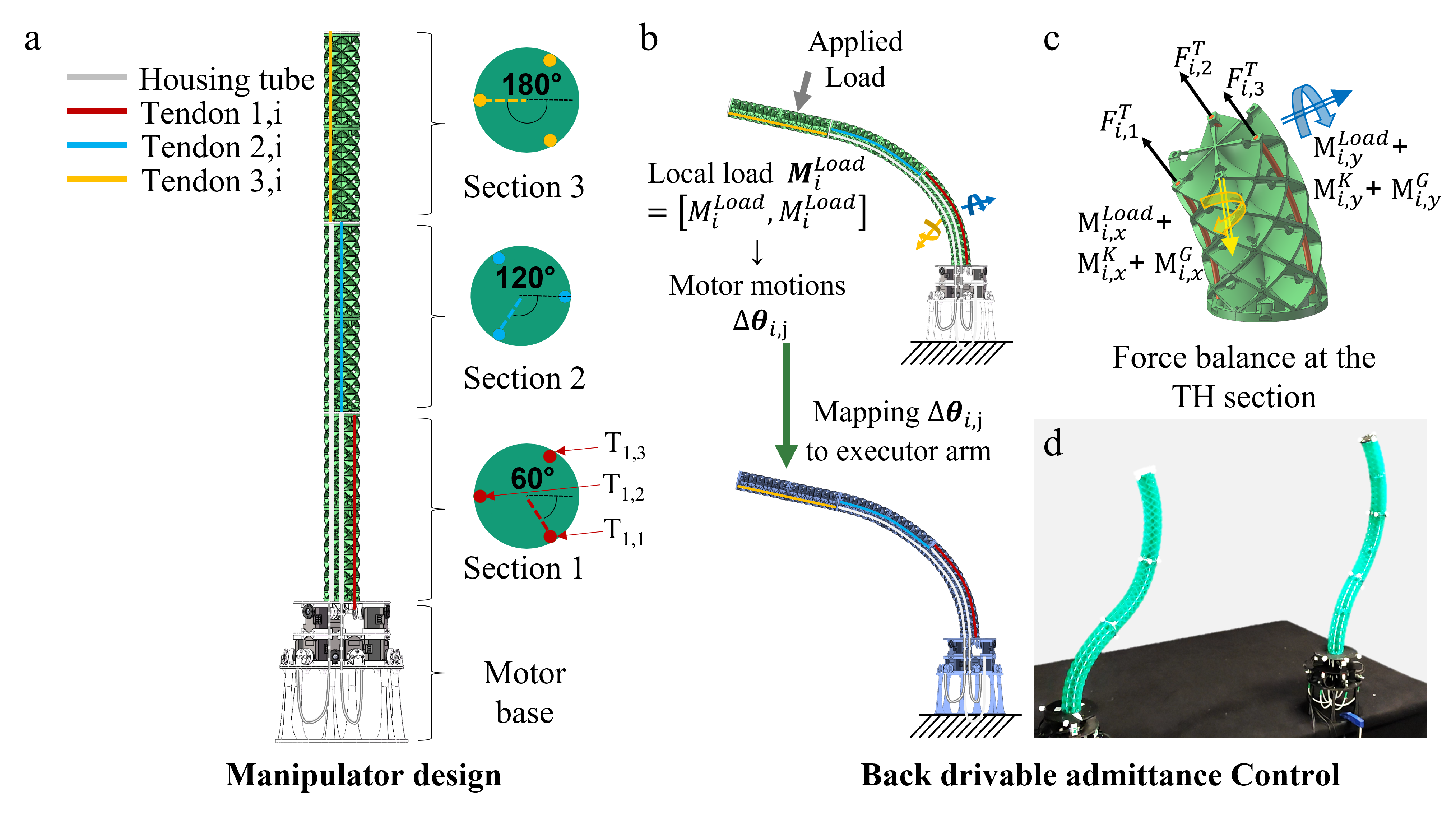}
\captionsetup{justification=centering}
\caption{Soft twin arm Design and mechanism}
\label{fig: Manipulator Design}
\end{figure*}

The demonstrator arm must allow the configuration to be easily altered by external forces, whilst holding and estimating its pose. 
By combining the TH arm structures with back-drivable servo motors enables the entire arm configuration to adapt to externally applied load, as shown in Fig.~\ref{fig: Manipulator Design}d. When the operator ceases motion and releases the applied force, the local load returns to zero. However, the dynamic friction within the tendons and motors transitions into static friction(Eq.~\ref{Eq: Friction}), which assists the Trimmed Helicoid (TH) segment in counterbalancing its initial body weight and elastic restoring force under any given configuration(Fig.~\ref{fig: Manipulator Design}c and Eq.~\ref{Eq: Force balance}). As a result, the robotic arm is capable of maintaining its configuration even after the completion of the interaction-based teaching process, as shown in Fig.~\ref{fig: Manipulator Design}d.   

The force equilibrium governing the tendon-driven actuation of the soft robotic arm is expressed as:

\begin{equation}
    \begin{split}        
    &\mathbf{F}^{T}_i+\mathbf{F}^{act}_i+\mathbf{F}_{f,i}=0\\
        &\mathbf{F}^{act}_i=k_{act}\mathbf{I_i}+C_{act}.      
    \end{split}
    \label{Eq: Friction}
\end{equation}

\noindent Here, \(\mathbf{F}^{T}_i \in \mathbb{R}^3\) represents the tendon tension force in section \(i\), responsible for driving the soft robotic module. The actuation force generated by the servo motors is denoted as \(\mathbf{F}^{act}_i \in \mathbb{R}^3\), with \(\mathbf{I}_i \in \mathbb{R}^3\) representing the corresponding input currents. The relationship between the actuation force and the input current is characterized by the coefficients \(k_{act}\) and \(C_{act}\).

The cumulative frictional force \(\mathbf{F}_{f,i} \in \mathbb{R}^3\) arises from contact between the tendon housing tubes and the interaction between tendons and tendon housing (TH) structures. This force is modeled based on two distinct frictional states:

\noindent 1. Static Friction (\(\dot{\mathbf{x}}_i = 0\)):  
   When the tendon is stationary, the friction force remains within the static friction threshold, expressed as:
   \begin{equation}
   -\mathbf{F}_{s,i}^{max} < \mathbf{F}_{s,i} < \mathbf{F}_{s,i}^{max},
   \end{equation}
   where the maximum static friction force is given by:
   \begin{equation}
   \mathbf{F}_{s,i}^{max} = \mu_s \left(\alpha \mathbf{F}^{T}_i + \beta \mathbf{F}^{act}_i \right).
   \end{equation}
   Here, \(\mu_s\) represents the static friction coefficient, while \(\alpha\) and \(\beta\) denote the respective contributions of the tendon tension force and the actuation force to the cumulative friction. These coefficients are assumed to be 0.5.

\noindent 2. Kinetic Friction (\(\dot{\mathbf{x}}_i \neq 0\)):  
   Once the tendon begins to move, kinetic friction governs the resistance, defined as:
   \begin{equation}
   \mathbf{F}_{k,i} = \mu_k \left(\mathbf{F}^{T}_i + \mathbf{F}^{act}_i\right) \frac{{\dot{\mathbf{x}}}_i}{\left|{\dot{\mathbf{x}}}_i\right|}.
   \end{equation}
   The kinetic friction force can also be reformulated as:
   \begin{equation}
   \mathbf{F}_{k,i} = \left(k_{kf} \mathbf{I}_i + C_{kf} \right) \frac{{\dot{\mathbf{x}}}_i}{\left|{\dot{\mathbf{x}}}_i\right|},
   \end{equation}
   where \( \mu_k \) is the kinetic friction coefficient, and \( k_{kf} \) and \( C_{kf} \) characterize the dependence of kinetic friction on the actuation input.

By incorporating force equilibrium with frictional effects, the model provides a comprehensive representation of the mechanical interactions within the soft robotic arm. Then, the equilibrium conditions of the TH bending module can be described by the following equations:
\begin{equation}     
    \begin{split}
    \mathbf{M}^{T}_i = &F^T_{i,j}[sin(\phi_{i.j})R_T,~ cos(\phi_{i.j})R_T]^T,~j=1,2,3\\
    &\mathbf{M}^{T}_i+\mathbf{M}^{load}_i+\mathbf{M}^{G}_i+\mathbf{M}^{K}_i=0    
    \end{split}
    \label{Eq: Force balance}
\end{equation}
\noindent where the $\mathbf{M}^{T}_i$ is the resultant bending moment of tendon forces ($\mathbf{F}^{T}_i= F^T_{i,j}$) at the local frame attached to a specific point along the central line of section $i$. $\mathbf{\phi}_{i}=\phi_{i,j}$ depicts azimuth angles of the three tendons. Similarly, $\mathbf{M}^{load}_i$, $\mathbf{M}^{G}_i$, and $\mathbf{M}^{K}_i$ represent the local moments resulting from external load, initial gravity, and restoring forces.

To some extent, this methodology aligns with the fundamental principles of admittance control in rigid robots. However, for this soft and lightweight arm, admittance control is not achieved through active compensation of gravity and friction using complex dynamic models.  
Instead, it relies on the robots embodiment, which balances gravitational forces, frictional properties, and elastic restoring forces.

To be able to control the second arm we must also be able to perceive the robots pose of the physical twin. By maintaining continuous tendon tension via current-controlled motors, the robot inherently perceives its configuration and interaction forces via the tendon length.  The tendon lengths describe the pose of the robot and can be used to directly set the configuration of the executer robotic arms.

\subsection{Mapping to Executer Robot}

The use of Physical Twin Control (PTC) framework enables direct mapping between a physical twin demonstrator and a soft robotic executor, by leveraging the tendon-driven actuation. This mapping can be applied not only in 1:1 direct mapping but also 1:X scaled mapping.

In the 1:1 tendon mapping, the demonstrator and executor arms share identical structural and actuation properties. Each tendon in the demonstrator directly corresponds to a tendon in the executor, ensuring full deformation replication through precise mapping The demonstrator’s physical interactions, are passively transferred to the soft robotic executor with high fidelity. This approach allows for intuitive and real-time teleoperation, leveraging the compliance and natural deformation of the physical twin (Fig.~\ref{fig: Concept physical twin}b).

Beyond identical mapping, a 1:X mapping extends this control framework to scaled-up or scaled-down executor robots (Fig.~\ref{fig: Concept physical twin}c). In this configuration, the tendon-driven deformation of the demonstrator (small arm) is proportionally scaled to a larger or smaller executor arm. The scaling factor \(X\) dictates the proportional change in tendon displacement and arm curvature, ensuring consistent deformation patterns across different sizes. The use of a 1:X mapping enables a human scale demonstrator to be control a manipulator with a larger workspace, increasing the reachability.  Conversely, a scaled-down executor enables more precise manipulation for small-scale tasks, such as delicate object handling in constrained environments.


\subsection{Executer Robot Arm Control: Active friction and stiffness control}

The physical twin demonstrator detects interaction loads by measuring tendon length variations caused by passive deformation and adapts to load direction by reconfiguring its overall shape. However, in a multi-degree-of-freedom (Multi-DoF) continuum arm, multiple solutions may exist under identical conditions. To introduce alternative configurations, adjustable friction and stiffness properties were incorporated, allowing the soft arm to exhibit distinct responses to the same external stimulus.  

Since the static friction threshold $\mathbf{F}_{s,i}^{max}$ and kinematic friction force are positively correlated with the actuation force of tendons, increasing the total or average input current ($\mathbf{F}_i, I_{av,i}$) enhances both dynamic damping and static holding capabilities. Moreover, by regulating the input current $\mathbf{I}_{i}$ individually to proportionally adjust the resultant moment output $\mathbf{M}^{T}_i$ in response to bending deformation, each section of the arm gains additional adjustable stiffness.

\begin{figure}[htb]
\centering
\includegraphics[width=0.5\textwidth]{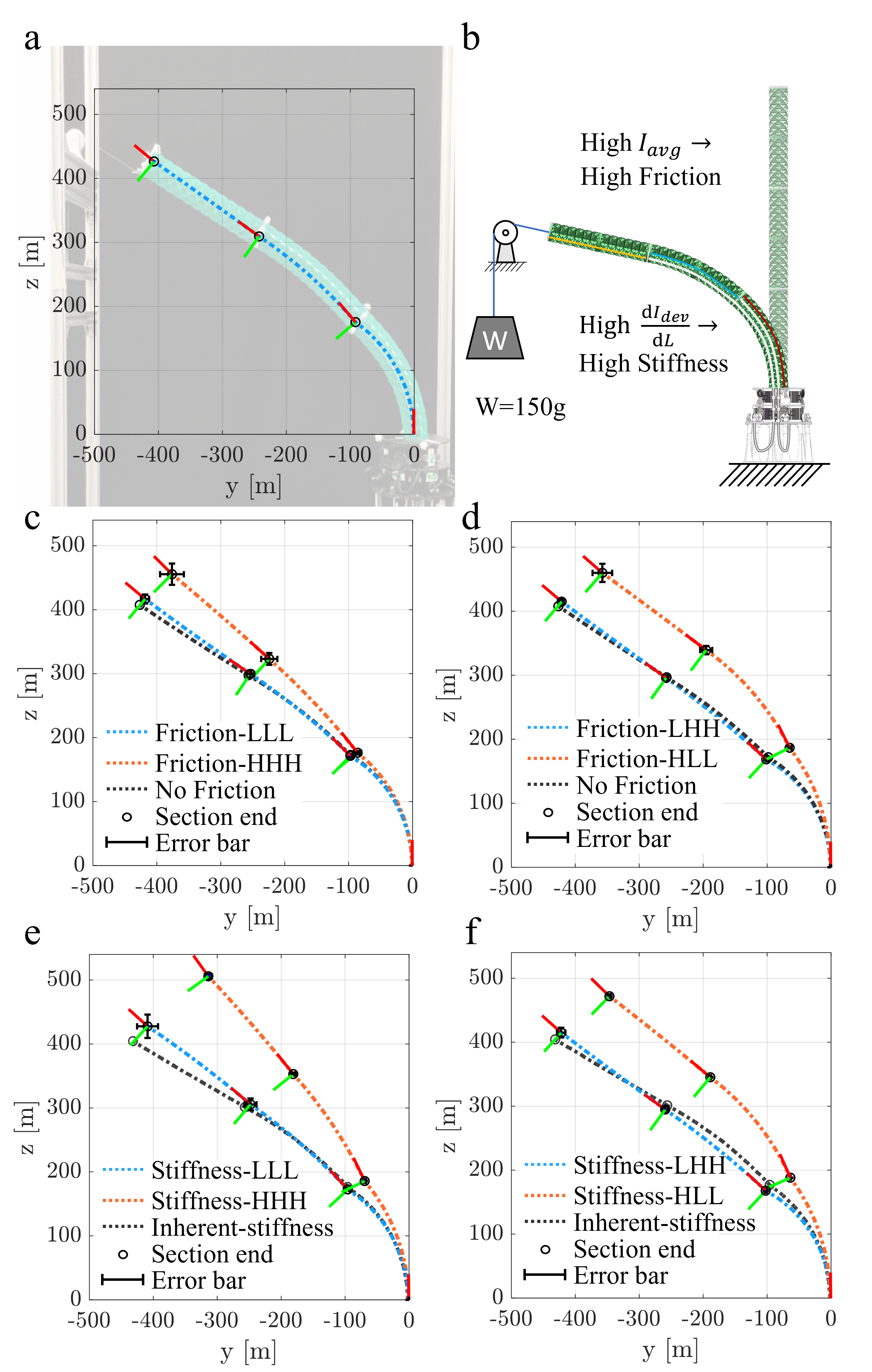} 
\caption{Influence of Friction and Stiffness Regulation on Deformation Response.
(a) Initial Deformation Under External Load.
(b) Experimental Setup for Load Application.
(c) Effect of Friction Distribution (LLL vs. HHH).
(d) Effect of Friction Distribution (LHH vs. HLL).
(e) Effect of Stiffness Variation (LLL vs. HHH).
(f) Localized Stiffness Regulation (LHH vs. HLL). }
\label{fig: Friction and stiffness regulation}
\end{figure}

Figure~\ref{fig: Friction and stiffness regulation}a illustrates the initial deformation of the robotic arm under an externally applied load, generated by the experimental setup depicted in Figure~\ref{fig: Friction and stiffness regulation}b. By adjusting the friction and stiffness distribution across Sections I–III, the resulting shape configuration can be modulated (Figure~\ref{fig: Friction and stiffness regulation}c–f). 
As shown in Figures~\ref{fig: Friction and stiffness regulation}c–d, when Section I (near the base) exhibits low friction with an LLL (Low-Low-Low) or LHH (Low-Low-High) distribution, the arm closely follows its initial deformation in the absence of current input and with slack tendons. Conversely, when friction is distributed as HHH (High-High-High) or HLL (High-Low-Low), resulting in increased friction at the base, the robotic arm stabilizes in a less deformed configuration, particularly in Section I.
Figures~\ref{fig: Friction and stiffness regulation}e–f illustrate the influence of adjustable stiffness on deformation response under external loading. A comparison between LLL and HHH stiffness distributions indicates that higher overall stiffness constrains deformation, whereas a more compliant arm exhibits greater deflection. The LHH configuration enhances bending flexibility in Section I and increases displacement in Sections II and III. In contrast, the HLL configuration restricts deformation at the base, leading to higher impedance under load.
 
These findings align with previous observations (Fig.~\ref{fig: Friction and stiffness regulation}c-d), where the combination of friction and stiffness regulation enables the arm to adopt different equilibrium configurations. By tuning actuation force-dependent static friction thresholds and stiffness modulation, the system can enhance static holding capabilities and provide more alternative response to external stimulus, enhancing adaptability for multi-degree-of-freedom soft robotic arms.

\section{Experimental Setup}

The main structure of the robotic arm was manufactured via 3D printing with thermoplastic polyurethane (TPU) filament (98A, Purefil). The robot arm has a length of 60 cm, while the larger white arm extends to 98 cm. The base section of the larger white arm, along with its demonstrator, is designed with a conical shape featuring a taper angle of 14°. Two of the smaller demonstrator arms were actuated by servomotors (Dynamixel XL330-M077-T, ROBOTIS INC). The smaller green executor arm was driven by a XC330-T288-T servo motor. The larger white executor arm was activated with a combination of Dynamixel XL430-W250 servomotors and Feetech 45BL motors.

For all these arms, the first section (of 3) is directly connected to motors via pulley wheels, whereas Sections II and III utilize housing tubes to guide the cables through pre-designed channels within the TH structure. The housing tube-cable system operates similarly to a bicycle brake system, transmitting cable displacement without deforming the outer tube. The tubes are fixed at both ends, ensuring controlled force transmission from the motor base to specific sections of the arm.

\begin{figure}[htp]
\centering
\includegraphics[width=3in]{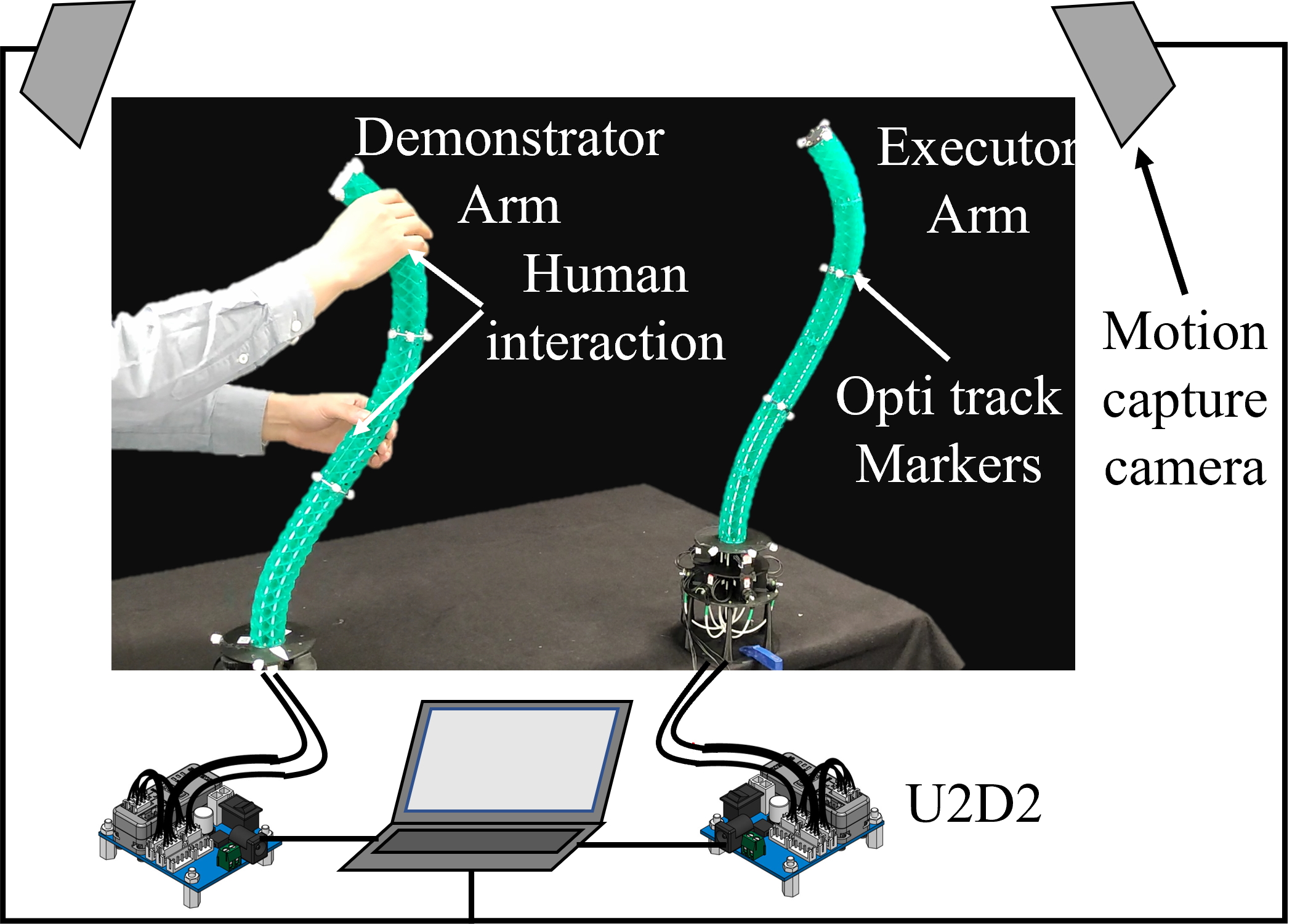}
\caption{Experimental Setup for teleportation and data collection of twin arm}
\label{fig: Experiment set up}
\end{figure}

These arms were regulated utilizing the U2D2 (DYNAMIXEL) / FE-URT-1 (Feetech) interface and their accompanying software development kit (SDK). Communication between the two arms was facilitated through a PC running MATLAB. The workspace and motion of the robotic arms were recorded and analyzed using a Motion Capture System (OptiTrack), as shown in Fig.~\ref{fig: Experiment set up}.
 
\section{Experimental Results}

\subsection{Motion Mapping Between the Twin Arms}

\begin{figure}[htp]
\centering
\includegraphics[width=0.5\textwidth]{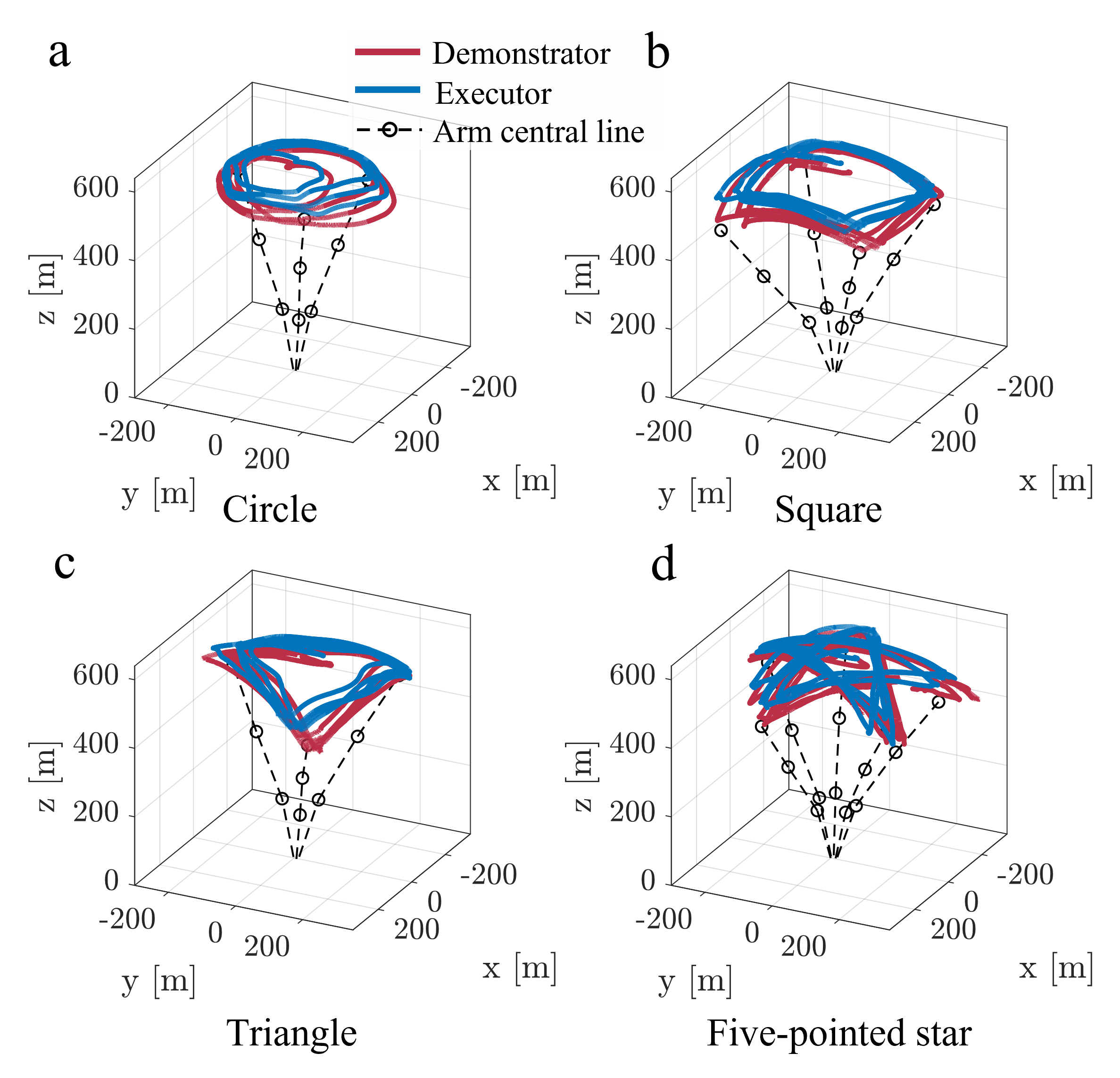}
\caption{Teleoperationed trajectories of the Demonstrator and Executor Arms. (a) Circular trajectory, (b) Square trajectory, (c) Triangular trajectory, (d) Five-pointed star trajectory.}
\label{fig: Teleopration Circles}
\end{figure}

We first characterize the capacity to map the motion between the two arms.  Fig.~\ref{fig: Teleopration Circles} presents the three-dimensional trajectories of the demonstrator (red) and executor (blue) arms for different predefined motion patterns input through the interaction with the demonstrator arm: circular, square, triangular, and five-pointed star trajectories.  These trajectories illustrate the executor's ability to replicate the demonstrator's motion, with variations reflecting system dynamics, compliance, and teleoperation fidelity.

As shown in Table~\ref{tab:deviation}, the relative deviation between the demonstrator and executor arms varies across different trajectory shapes. It is observed that as the trajectory becomes sharper (e.g., transitioning from a circular to a five-pointed star pattern), the deviation in the x- and y-axes decreases, while the deviation in the z-axis increases. This trend suggests that sharper trajectories improve in-plane accuracy while leading to greater vertical displacement errors. The reduction in x- and y-axis deviations may be attributed to the reduced influence of hysteresis and material relaxation effects when the deformation direction changes more frequently and significantly. Conversely, the increased error in the z-axis may result from higher dynamic accelerations, which introduces additional inertial loads and elevated tendon forces, affecting vertical stability.

\begin{table}[h]
    \centering
    \caption{Relative Deviation Between the Demonstrator and Executor Arms of Different Trajectories}
    \renewcommand{\arraystretch}{1.2} 
    \begin{tabular}{|l|c|c|c|}
        \hline
        \textbf{Shape} & \textbf{x(\%)} & \textbf{y(\%)} & \textbf{z(\%)} \\ 
        \hline
        Circle ($\bigcirc$) & 7.74 & 9.04 & 1.28  \\ 
        \hline
        Square ($\square$) & 8.10 & 8.89 & 1.82  \\ 
        \hline
        Triangle ($\triangle$) & 7.80 & 8.73 & 2.63  \\ 
        \hline
        Five-pointed star ($\star$) & 7.21 & 7.34 & 3.40  \\ 
        \hline
    \end{tabular}
    \label{tab:deviation}
\end{table}

\subsection{Directional Error of the Mapping}

\begin{figure}[ht]
\centering
\includegraphics[width=0.5\textwidth]{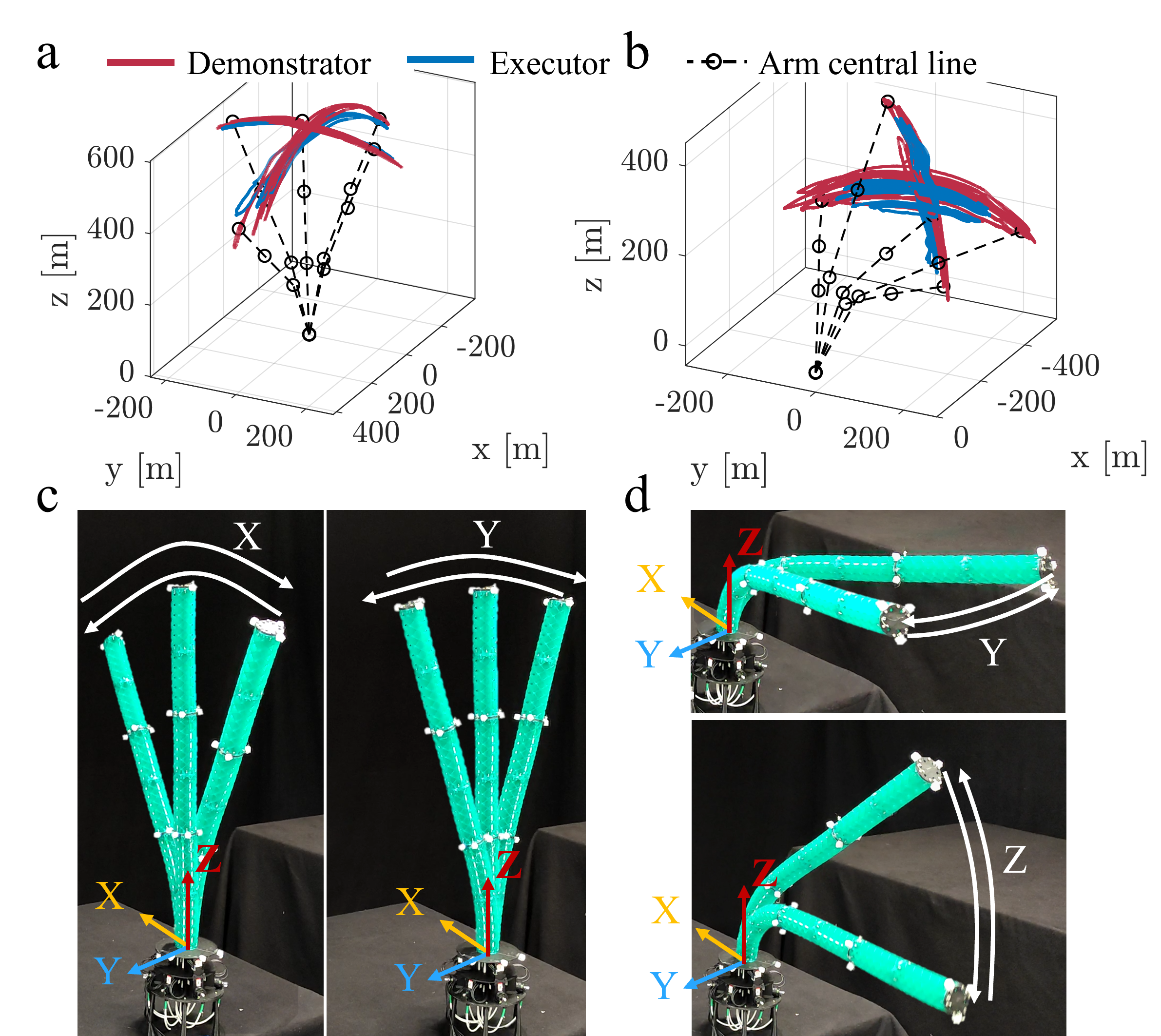}
\caption{Teleoperation Performance Across Different Configurations and Motion Directions. (a) \& (b) 3D tracking trajectories of the demonstrator and executor in vertical and horizontal configurations.
(c) Vertical configuration: Motion along the X- and Y-axes.
(d) Horizontal configuration: Motion along the Y- and Z-axes.}
\label{fig: Teleopration Error diff config dirct}
\end{figure}

The robot arms have a directionality, and this affects the mapping error across their workspace with different direction showing different performance.
Fig.~\ref{fig: Teleopration Error diff config dirct} shows the teleoperation performance of the demonstrator (red) and executor (blue) robotic arms in a vertical and horizontal configuration with different movement directions (moving in the X, Y, and Z axes). The top row shows three-dimensional trajectory tracking results, where the executor replicates the demonstrator's motion, with variations reflecting system compliance and control accuracy. The bottom row provides experimental snapshots highlighting distinct movement directions.

The relative error between the demonstrator and executor arms, is also quantified in Table~\ref{tab:rel_deviation}. 
These results show that the vertical configuration movements exhibit the highest error along the primary movement direction, with Y-axis motion showing the largest deviation in the Y-direction (7.97\%), and X-axis motion in the X-direction (9.91\%). In contrast, horizontal configuration movements result in larger deviations, particularly along the Y-axis (15.25\%) for Y-direction motion and along the Z-axis (7.88\%) for Z-direction movement. This suggests that horizontal configurations introduce more significant lateral deviations, whereas vertical configurations primarily affect motion along the intended direction. The observed variations are likely due to system compliance, gravitational influences, and differences in tendon force distribution across configurations. This suggests that optimizing the arm’s configuration can effectively minimize the deviation between the demonstrator and executor.

\begin{table}[h]
    \centering
    \caption{Relative error between the demonstrator and executor.}
    \renewcommand{\arraystretch}{1.2} 
    \begin{tabular}{|c|c|c|c|c|}
        \hline
        \multirow{2}{*}{\textbf{Configuration}} & \multirow{2}{*}{\textbf{Moving Direction}} & \multicolumn{3}{c|}{\textbf{Position Axis}} \\ 
        \cline{3-5}
        & & \textbf{X(\%)} & \textbf{Y(\%)} & \textbf{Z(\%)} \\ 
        \hline
        \multirow{2}{*}{Vertical} & Y & 1.35 & 7.97 & 1.57 \\ 
        \cline{2-5} 
        & X & 9.91 & 3.46 & 3.76 \\ 
        \hline
        \multirow{2}{*}{Horizontal} & Y & 3.33 & 15.25 & 2.83 \\
        \cline{2-5} 
        & Z & 3.72 & 2.71 & 7.88 \\ 
        \hline
    \end{tabular}
    \label{tab:rel_deviation}
\end{table}

\subsection{Human-robot Interactions Under Different Stiffness Distributions}

\begin{figure}[htp]
\centering
\includegraphics[width=0.5\textwidth]{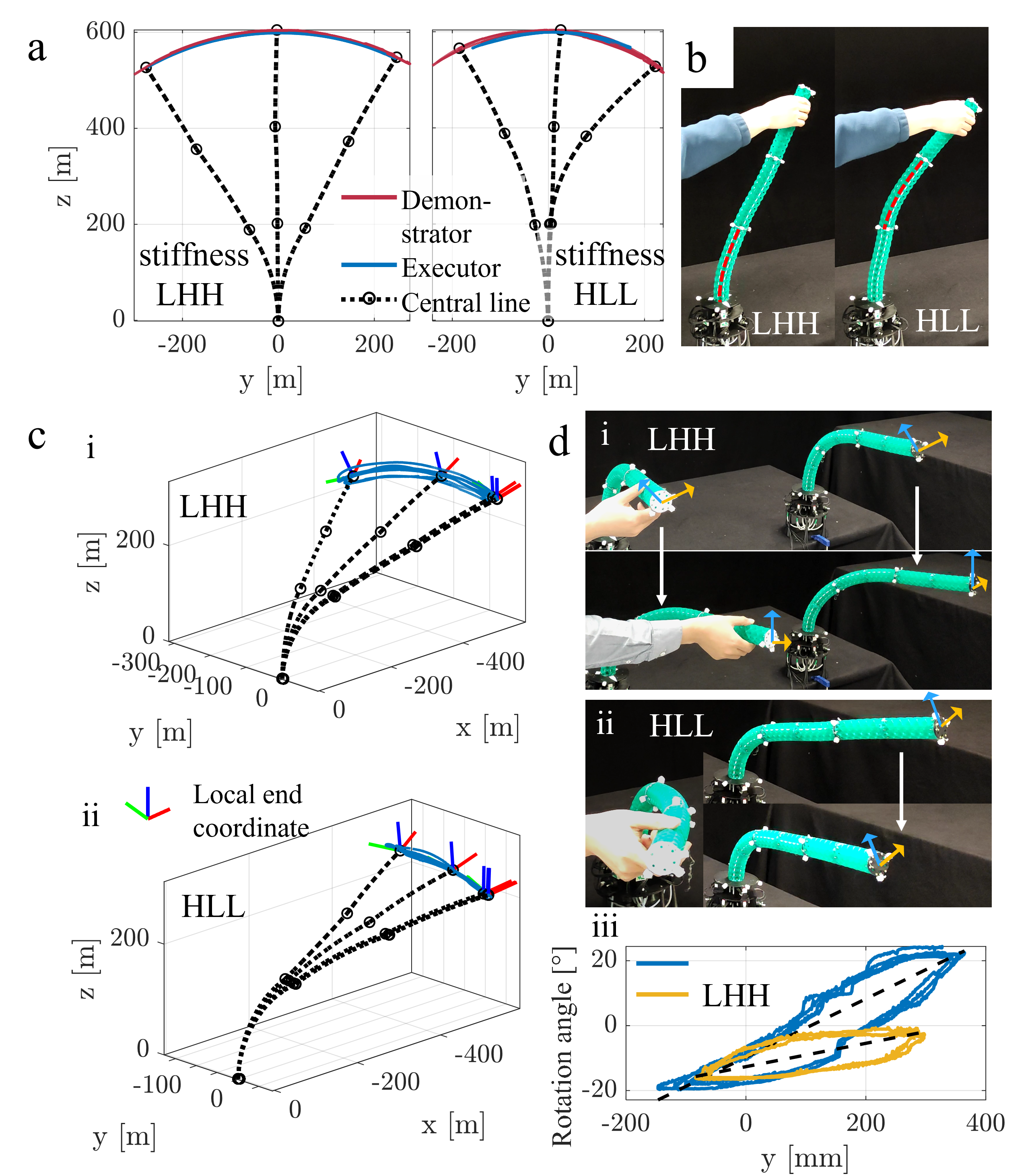}
\caption{Effect of Stiffness Distribution (LHH vs. HLL) on Teleoperation Performance. (a) Interaction trajectories with vertical configuration.
(b) Interaction Response with vertical configuration.
(c) Interaction trajectories with horizontal configuration. 
(d) End-Effector Rotation and Translation with horizontal configuration.}
\label{fig: Interaction with different stiffness}
\end{figure}

As described in Section II.C, the stiffness properties of the demonstrator robot can be varied.  In the following experiments we investigate how different stiffness distributions affect the teleoperation accurarcy and the interactions of the executor robot when in a vertical configuration.
Fig.~\ref{fig: Interaction with different stiffness}a\&b  compares two different stiffness distributions: LHH (Low-High-High) and HLL (High-Low-Low). 
In the LHH configuration, the relative deviations along the X-, Y-, and Z-axes are 1.2\%, 8.7\%, and 2.1\%, respectively. In contrast, the HLL configuration leads to  higher errors, particularly in the Y- and Z-directions, with values of 1.2\%, 12.0\%, and 3.7\%, respectively. These findings indicate that a stiffer base (HLL) results in increased vertical deviation and reduced compliance during interaction, likely due to the higher interaction forces induced by the stiffness distribution.

Fig.~\ref{fig: Interaction with different stiffness}c\&d compare translation range and translation-rotation coupling effect with the horizontal configuration under the two stiffness distributions (LHH and HLL). Fig.~\ref{fig: Interaction with different stiffness}d-iii shows the relationship between the arm end rotation angle and the Y axis translation, revealing larger translation range with LHH stiffness distribution and less translation coupling with HLL stiffness distribution under horizontal configuration.

These findings suggest that motion deviation and coupling effects can be mitigated by optimizing the stiffness distribution, thereby enhancing human operation and mapping performance. However, further investigation is required to fully understand the underlying working principles and extend their applicability to more complex tasks.

\subsection{Demonstration of Teleoperation}

\subsubsection{Navigation Through a Narrow Gap}
\begin{figure*}[htp]
\centering
\includegraphics[width=1\textwidth]{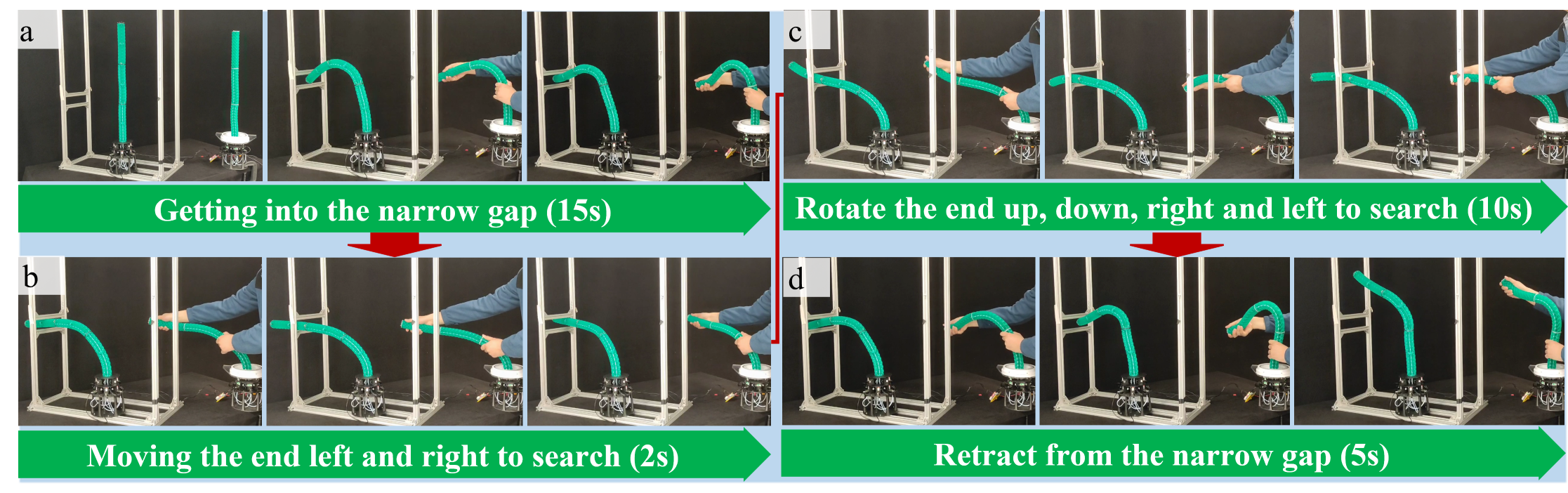}
\caption{Teleoperated Manipulation for Exploration Through a Narrow Gap: (a) Entry Phase, (b) Lateral Search, (c) Rotational Search, and (d) Retraction Phase.}
\label{fig: through narrow gap}
\end{figure*}

We use a challenging navigation scenario to showcase how the intrinsic compliance and structural similarity between the soft twin arms enables intuitive control and interaction of multi-degree-of-freedom (Multi-DoF) continuum robotic systems (Fig.~\ref{fig: Concept physical twin}b). This task relies upon  exploitation of the robots compliance and the ability to control its full configuration.  Fig.~\ref{fig: through narrow gap} illustrates the teleoperated maneuvering of a flexible robotic arm through a constrained environment, structured into four key phases:

\begin{itemize}
    \item \textbf{Entry Phase (15s)}: The arm is carefully guided into the narrow gap, leveraging its compliance to ensure smooth insertion without excessive resistance.
    \item \textbf{Lateral Search (2s)}: The end-effector moves horizontally to scan the surroundings for spatial awareness.
    \item \textbf{Rotational Search (10s)}: The end-effector undergoes controlled rotations in multiple directions (up, down, left, and right) for enhanced environmental exploration.
    \item \textbf{Retraction Phase (5s)}: The arm is safely withdrawn, minimizing potential collisions or structural interference.
\end{itemize}

To enhance maneuverability, an adaptive stiffness modulation strategy is employed. Initially, the LLL stiffness configuration is applied, facilitating large-range deformations for seamless insertion. Once inside the gap, the LHH stiffness mode is activated, enabling effective lateral scanning. For rotational exploration, the HLL stiffness configuration is utilized, allowing decoupled up-down and left-right rotations for detailed observation. Finally, the LLL stiffness mode is re-engaged to ensure smooth and effortless retraction.

This sequential stiffness adjustment optimizes both compliance and control precision, allowing efficient navigation through constrained environments while maintaining stability and adaptability.

\subsubsection{Workspace Expansion with a Scaled-Up Twin}

\begin{figure}[htp]
\centering
\includegraphics[width=0.5\textwidth]{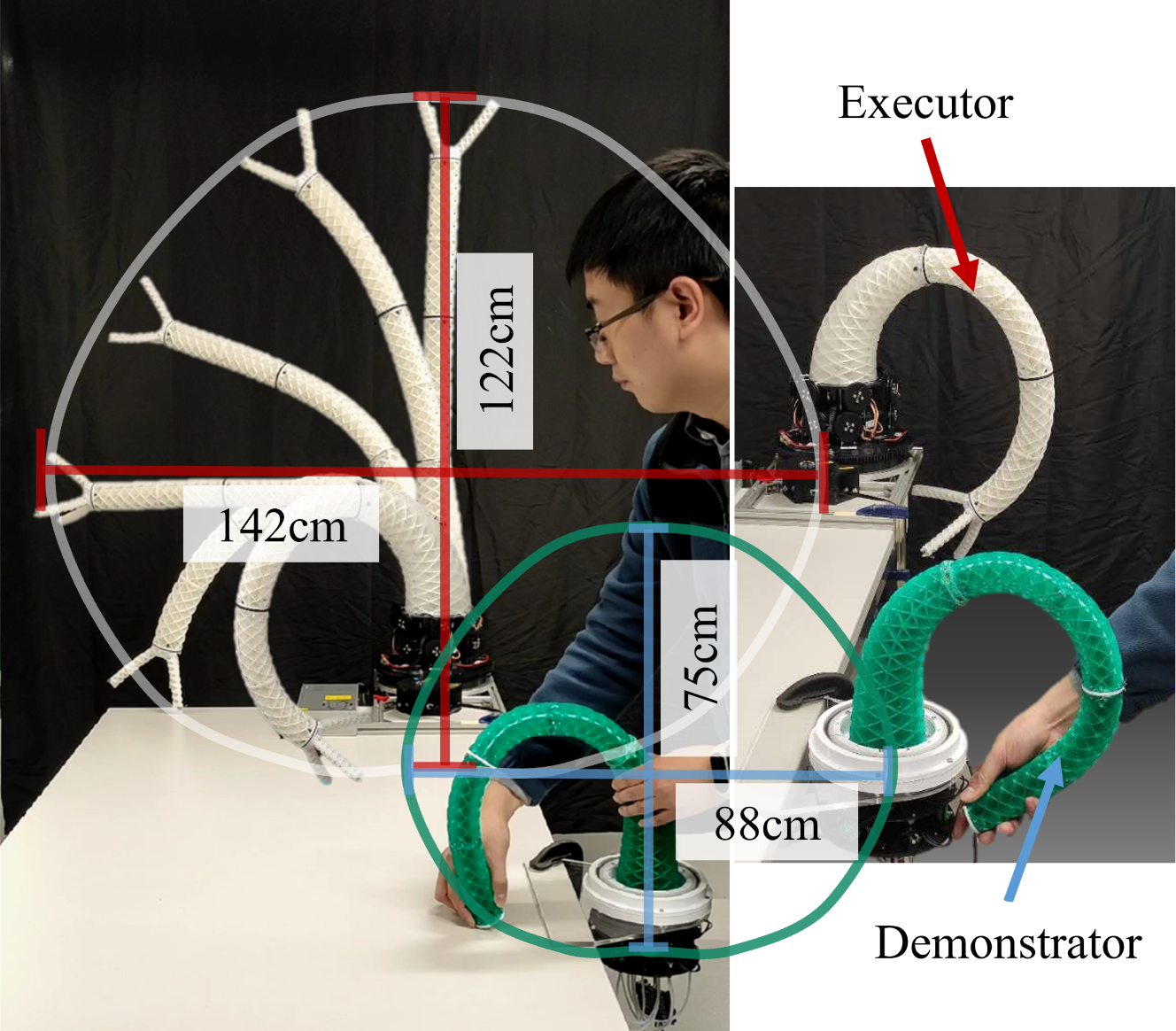}
\caption{Workspace Expansion via Scaled-Up Executor Arm: Comparison of motion range between the demonstrator and the larger executor.}
\label{fig: expanded work space}
\end{figure}

Fig.~\ref{fig: expanded work space} demonstrates the expansion of the operational workspace achieved through a scaled-up executor arm in comparison to the demonstrator arm. The demonstrator arm (green, length: 61 cm) is controlled by a human operator, while the executor arm (white, length: 98 cm) with a three-finger gripper (Fig.~\ref{fig: Concept physical twin}b and Fig.~\ref{fig: Concept physical twin})~\cite{QinghuaDext2025} mirrors these movements at a larger scale, significantly extending the workspace. 

Quantitatively, the executor achieves a workspace expansion of 142 cm in width and 122 cm in height, compared to the demonstrator's 88 cm width and 75 cm height, as shown in Fig.~\ref{fig: expanded work space} . This scaling approach effectively extends the robot’s reach, enabling larger motion execution while preserving the fidelity of teleoperation.

By maintaining compliance and adaptability, the scaled-up executor enhances reachability and manipulation efficiency, particularly for applications involving remote operations, large-scale inspections, and interaction within extended environments. This approach ensures that fine motor control from the demonstrator is precisely translated into large-scale movements, maintaining maneuverability without sacrificing precision.

\section{Conclusion and Discussion}

This work introduced the Physical Twin Control (PTC) framework, which employs a physical twin as a real-world simulator for direct teleoperation of a tendon-driven soft robotic arm. By maintaining an identical structure and configuration space, the physical twin inherently captures its full deformation state, enabling high-fidelity, real-time configuration mapping with minimal computational effort. This eliminates the need for complex kinematic models while preserving intuitive, global control of the entire soft arm. Operators can seamlessly transfer complex, compliant motions by leveraging the twin’s natural deformation, and by tuning its stiffness and friction properties, localized control can be achieved, enhancing adaptability in dynamic environments.

While the PTC framework offers intuitive and efficient control, several limitations remain. Precision control for fine manipulation tasks requires further optimization, and the reliance on direct teleoperation limits autonomy. Future work should explore hybrid integration with machine learning-based predictive models to improve automation and intelligent motion planning. Additionally, incorporating multi-modal sensory feedback (e.g., force, vision, and haptics) could refine perception and control accuracy~\cite{pan2024vision, chen2024data}.

Future research will prioritize enhancing scalability and adaptability, enabling seamless interaction across varying robotic sizes and configurations. The integration of data-driven learning mechanisms could facilitate learning from teleoperated demonstrations, thereby reducing dependency on real-time human input and enhancing predictive capabilities. By further advancing the Physical Twin Control (PTC) paradigm, this framework has the potential to establish efficient, intuitive, and scalable control strategies for soft robotic arms. Additionally, it offers a promising approach for data collection, contributing to the development of generative controllers that enhance autonomous decision-making and adaptability in soft robotics.






\bibliographystyle{IEEEtran}
\bibliography{References}


\end{document}